 \documentclass[pmlr,twocolumn,10pt]{jmlr} 

\usepackage{booktabs}
\usepackage{float}
\usepackage{siunitx}

\newcommand{\equal}[1]{{\hypersetup{linkcolor=black}\thanks{#1}}}

\jmlrvolume{225}
\jmlryear{2023}
\jmlrsubmitted{LEAVE UNSET}
\jmlrpublished{LEAVE UNSET}
\jmlrworkshop{Machine Learning for Health (ML4H) 2023} 

 \title[Visual Classification as Linear Combination of Words]{Representing visual classification as a linear combination of words}

\author{%
\Name{Shobhit Agarwal} \Email{shobhitagarwal122@gmail.com}\\
\addr Dana-Farber Cancer Institute, Boston, MA \\
\addr Reedy High School, Frisco, TX
\AND
\Name{Yevgeniy R. Semenov} \Email{ysemenov@mgh.harvard.edu}\\
\addr Massachusetts General Hospital, Boston, MA \\
\addr Harvard Medical School, Boston, MA
\AND
\Name{William Lotter}\equal{Corresponding author} \Email{lotterb@ds.dfci.harvard.edu}\\
\addr Dana-Farber Cancer Institute, Boston, MA \\
\addr Brigham and Women's Hospital, Boston, MA \\ 
\addr Harvard Medical School, Boston, MA
}

\begin{document}

\maketitle

\begin{abstract}

Explainability is a longstanding challenge in deep learning, especially in high-stakes domains like healthcare. Common explainability methods highlight image regions that drive an AI model’s decision. Humans, however, heavily rely on language to convey explanations of not only ``where'' but “what”. Additionally, most explainability approaches focus on explaining individual AI predictions, rather than describing the features used by an AI model in general. The latter would be especially useful for model and dataset auditing, and potentially even knowledge generation as AI is increasingly being used in novel tasks. Here, we present an explainability strategy that uses a vision-language model to identify language-based descriptors of a visual classification task. By leveraging a pre-trained joint embedding space between images and text, our approach estimates a new classification task as a linear combination of words, resulting in a weight for each word that indicates its alignment with the vision-based classifier. We assess our approach using two medical imaging classification tasks, where we find that the resulting descriptors largely align with clinical knowledge despite a lack of domain-specific language training. However, our approach also identifies the potential for ‘shortcut connections’ in the public datasets used. Towards a functional measure of explainability, we perform a pilot reader study where we find that the AI-identified words can enable non-expert humans to perform a specialized medical task at a non-trivial level. Altogether, our results emphasize the potential of using multimodal foundational models to deliver intuitive, language-based explanations of visual tasks. 

\end{abstract}

\section{Introduction}
\label{sec:intro}

The `black box' nature of standard deep learning approaches is a long referenced challenge, where the need for explainability is especially emphasized in high-stakes domains like healthcare \citep{Van_der_Velden2022-cu, Gao2022-qv}. 
For computer vision-based applications, a common approach to explainability is the generation of `saliency maps' that are designed to highlight the image regions that drive a model’s decision. 
However, the utility and trustworthiness of saliency-based approaches have been questioned \citep{Arun2021-ds, Adebayo2018-uf}.
Additionally, saliency maps and similar localization-based techniques only address “where” and not “what”. 
While localization is a critical step in many applications, the lack of a more intuitive description of the features used by AI models can limit clinician trust and overall clinical integration.

Another understudied direction is the notion of task-level explainability. 
Instead of explaining the prediction for a single instance, a task-level (i.e., global) explanation would convey the general features used to make such predictions \citep{Kim2018-pn, Ghorbani2019-dt}. 
For example, an instance-level explanation would indicate why a particular lesion is classified as malignant, whereas a task-level explanation would convey differences between malignant and benign lesions in general.
Both levels of explanations have utility, but a task-level approach would be especially useful for purposes such as auditing and knowledge discovery. 
From an auditing perspective, methods are needed to efficiently identify spurious `shortcut connections' in AI datasets and the propensity of AI models to use these shortcuts \citep{Yan2023-gt, Kim2023-vt, Nauta2023-ph, Shirley_Wu2023-ac, Zhao2020-zc, De_Grave2023-sr}. 
Relatedly, there's a need for understanding if AI models are generally using similar features as clinicians or have identified new predictive features, especially as AI is increasingly applied to novel tasks.

Here, we present a strategy for more intuitive explainability that uses a vision-language model (VLM) to identify language-based descriptors of a visual classification task. 
Our approach is motivated by how humans heavily rely on language to describe differences between two visual categories, and can do so even for a new task without being explicitly told the features that distinguish the task.
Specifically, using a joint embedding space between images and text, our approach estimates a vision-based classifier as a linear combination of  word embeddings, resulting in a weight for each word that indicates its alignment with the visual task.
We specifically use CLIP \citep{CLIP} as the embedding space, a model that was trained in a contrastive fashion on publicly available image-caption data.
We assess feasibility in an extreme setting of identifying descriptive words for two medical imaging classification tasks without relying on domain-specific language training.
While domain-specific models can also be used, paired image-text data can be infeasible to collect for many medical tasks, and importantly, heavy supervision via human generated text may mask the use of shortcut connections or other predictive features not used by humans. 

The remainder of the paper is structured as follows. We first provide an overview of related work, followed by a description of our approach and analysis methods. We then present results across the two medical imaging datasets, including a human reader study to assess whether the AI-identified predictive words can guide non-experts to perform the tasks. We additionally explore the specific use case of dataset auditing to identify confounders. 
Code for our approach is available at: \url{https://github.com/lotterlab/task_word_explainability}.

\begin{figure*}[t]
\floatconts
  {fig:methods}
  {\caption{\vspace{-20pt}Modeling approach. Our approach uses a vision-language model with a joint embedding space between images and text (i.e., CLIP \citep{CLIP}) to represent a new visual classification task as a linear combination of words. The approach proceeds in two steps: 1) Fitting a linear classification model using image embeddings and classification labels, 2) Estimating the classifier as a linear combination of word embeddings, resulting in weights for each word that signify alignment with the visual task.}}
  {\includegraphics[width=\textwidth]{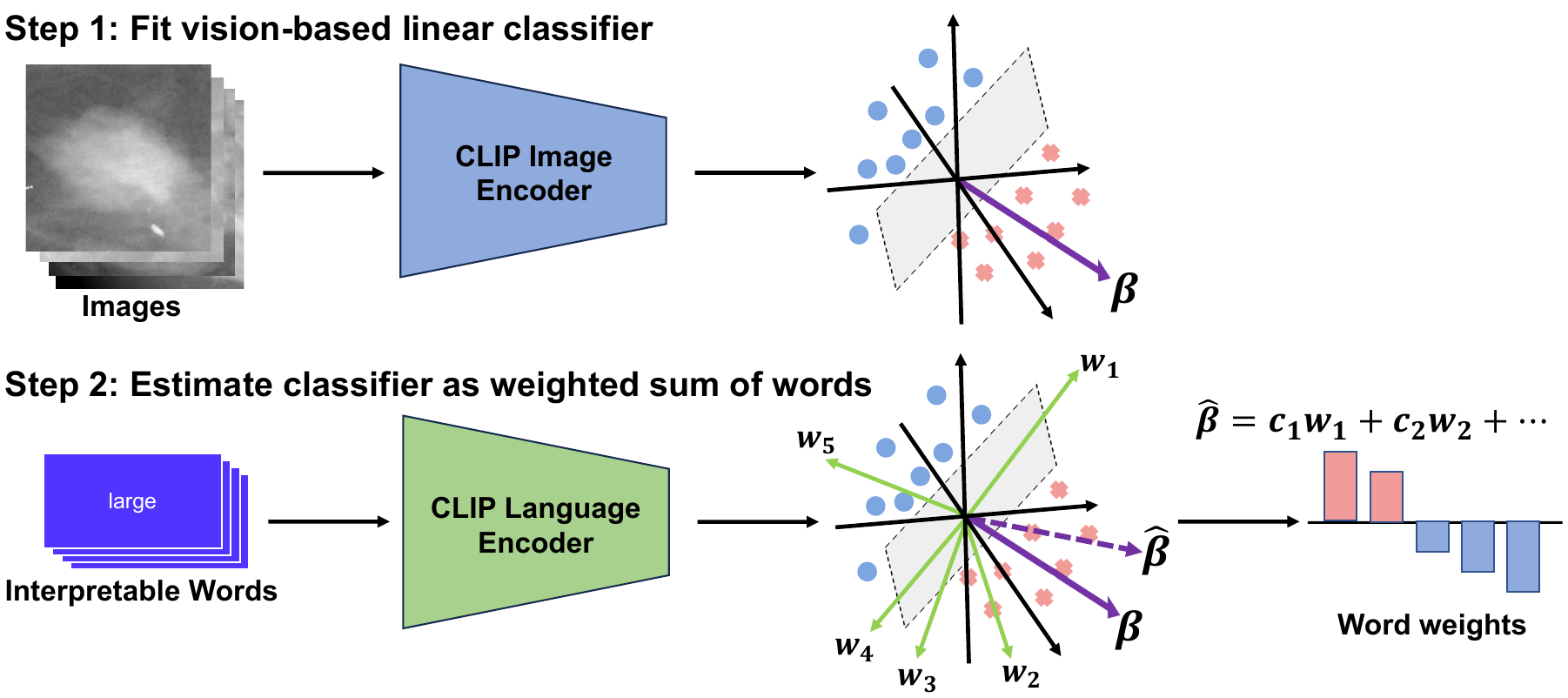}}
\end{figure*}

\section{Related Work}
\label{sec:related}

\subsection{Local vs. Global Explainability}
Explainability is a heavily studied topic in AI literature, with many different proposed approaches \citep{Van_der_Velden2022-cu, Gao2022-qv}.
Arguably, the most commonly applied approaches use local explanations, such as techniques like Grad-CAM \citep{Selvaraju2017-oj} that generate saliency maps to explain a model's prediction for a given instance.
An important line of work has also focused on global explanations that identify higher-level concepts across the dataset/model \citep{Ghorbani2019-dt}.
For instance, \citet{Kim2018-pn} introduced Concept Activation Vectors (CAVs) that are constructed from human-provided examples to represent concepts in a model's representation space, where a model's sensitivity to these concepts is then assessed using directional derivatives.
Recently, \citet{Yan2023-gt} presented a CAV-based approach to identify clinical and confounding concepts in dermoscopic images.
Efforts have also been made to combine local and global explanations \citep{Schrouff2021-bt, Achtibat2022-bj}, and to also use human-interpretable concepts to understand the features represented by individual neurons \citep{Bau2017-cr, Dani2023-bv, Schwettmann2023-yo, Goh2021-mg, bills2023, Hernandez2022-cr}. 
A common challenge across these methods is the heavy reliance on human supervision, such as by using human-selected examples to define concepts a-priori or by post-hoc human annotation of examples. 

\subsection{Natural Language Explanations}
Beyond saliency-based local explainability methods, recent efforts have aimed at generating natural language explanations for visual tasks \citep{Kayser2021-ja, Marasovic2020-gj, Wu2019-fs, Park2018-ao, Kim2018-nk, Hendricks2016-lw}. 
These approaches often involve a separate `explanation model' to generate language-based explanations of the original prediction model, where this separation between the prediction and explanation models creates risks of unfaithful explanations.
The recent approach by \citet{Sammani2022-ir} instead jointly formulates the task prediction and explanation as a text generation problem, where the model is trained using human-provided language explanations in a local explainability setting.

\subsection{Descriptor-based Visual Classification}
Along with the use of concepts and language for explainability, recent works have explicitly used language-based descriptors as a means to perform visual recognition tasks \citep{Pratt2022-rz, Yan2023-at, Maniparambil2023-ky, Lewis2023-gp, Liu2023-uy, Koh2020-ti, Yuksekgonul2023-fs}.
For instance, \citet{Menon2023-su} used GPT-3 \citep{gpt3} to generate text descriptors for each class in a dataset, e.g., ImageNet, and then used these descriptors as an alternative to the traditional zero-shot classification setting for CLIP where only class names are used for prompting \citep{CLIP}.
While descriptor-based visual classification is promising for robustness and improved instance-based interpretability, pre-defining descriptors for each class limits the identification of other predictive features or spurious correlations.

\section{Methods}
\label{sec:methods}

\subsection{Modeling Approach}
We sought to develop an approach for task-level, language-based explainability that does not require extensive human supervision or domain-specific language training. 
Our approach leverages a pre-trained, multimodal model with a joint embedding space for images and text. In this work, we specifically use CLIP with a ViT image encoder \citep{ViT} and a Transformer text encoder \citep{transformer}. 
CLIP was trained on a large amount of image-caption data to learn an aligned representation between images and their corresponding text captions \citep{CLIP}.
The learned feature space has been shown to be useful for downstream tasks, including relatively high linear-probe performance on several medical imaging datasets \citep{CLIP}.

With the pre-trained CLIP model, our approach proceeds in two steps (Figure 1). 
First, we train a linear classifier on top of the frozen embedding space using an image dataset with classification labels. 
Next we approximate the weights vector of the classifier as a linear combination of word embeddings. Doing so requires choosing a dictionary of words to be used. While a number of strategies could be used to select words (or even phrases), our goal here was to create a list of common words that could be used to describe visual classification tasks generally. 
As many words can have similar meanings, we also sought a sparse list of words to improve interpretability. 
We created such a list through a structured approach using ChatGPT (GPT 3.5 \citep{gpt3}). 
We first prompted ChatGPT for a list of general-purpose visual properties using the following prompt: “visual properties used to describe an object”. 
We then used the following prompt to generate adjectives for each of these properties: “a list of positive and negative adjectives for each of the properties listed above”. 
From this output, we selected representative words to result in the list in \tableref{tab:words_table}.

\begin{table}[htbp]
\floatconts
  {tab:words_table}%
  {\caption{List of general-purpose adjectives used.}}%
  {\vspace{-5pt}}
  {\begin{tabular}{l|ll}
  \bfseries Property & \bfseries Adjective 1 & \bfseries Adjective 2\\
  \hline
  Color & light & dark \\
  Shape & round & pointed \\
  Size & small & large \\
  Texture & smooth & coarse \\
  Transparency & transparent & opaque \\
  Symmetry & symmetric & asymmetric \\
  Contrast & low contrast & high contrast
  \end{tabular}}
\end{table}

Details of the two linear models trained in our approach are as follows. For the visual classification model, we use a logistic regression with a regularization parameter of $1$ and no intercept. For the word embedding model, we use a simple linear regression without an intercept to estimate the classification weight vector using the word embeddings as input features. 
This second regression results in a weight for each word in approximating the visual classifier. When trained with a binary label of 1 = “malignant” and 0 = “benign”, a positive weight for a word indicates alignment with a malignant prediction.

\subsection{Datasets}
We performed experiments using two datasets: 1) CBIS-DDSM \citep{cbis}, a collection of regions of interest (ROIs) of lesions identified on mammograms, and 2) the 2020 SIIM-ISIC Melanoma Classification Challenge dataset \citep{siim-isic}, a collection of dermoscopic images. For CBIS-DDSM, we consider a task of predicting whether a mass lesion is benign or malignant. For SIIM-ISIC, we consider a task of predicting whether a lesion is benign or malignant for melanoma. These tasks and datasets are detailed below.

CBIS-DDSM (Curated Breast Imaging Subset of DDSM) consists of scanned film mammograms with accompanying expert-drawn ROIs around lesions. Here, we specifically use the subset of ROIs corresponding to masses, which is the largest lesion subset in the dataset. We use the default training and testing splits in the dataset, which results in 1226 ROIs for training (617 malignant, 609 benign) and 365 for testing (145 malignant, 220 benign). 

The SIIM-ISIC dataset served as the basis for the 2020 SIIM-ISIC Melanoma
Classification Challenge and consists of dermoscopic images of histopathologically confirmed melanomas and benign melanoma mimickers.
As only the training set is publicly available for this dataset, we split this set into 80/20\% training/testing at the patient-level (i.e., no overlap in patients between training and testing).
We additionally subsample the benign images to achieve a more balanced 2:1 benign:malignant ratio, resulting in 1467 images for training (489 malignant, 978 benign) and 285 images for testing (95 malignant, 190 benign).

\subsection{Analysis Methods}
\subsubsection{Prototypical examples and shortcut analysis}
Prototypical example images were retrieved for the highest weighted words for each task. 
As each image may be correlated with several words, we specifically defined prototypes as examples that have a higher correlation with a particular word than what would be predicted from the other words alone.
In more detail, the dot product was first computed between each image in the training set and each word in the dictionary to measure the similarity between each pair.
A linear regression was then fit for each word to predict the dot product for that word for a given image based on the dot products of the remaining words.
The residual between the observed dot product and predicted dot product was then used as the prototype score, where a higher residual represents a higher prototype score.
A similar approach was taken to obtain prototypes for potential shortcut connections, where a word representing a potential shortcut was added to the original dictionary to obtain a weight for this word and also obtain prototypical examples. 
Code for computing the prototype scores is included in the Github repository.

\subsubsection{Pilot Reader Study}
We performed a reader study to more functionally assess the interpretability of our approach.
As argued in other works, human user studies can help go beyond qualitative metrics to assess human utility of explainability methods \citep{Poursabzi-Sangdeh2021-dl, Shen2020-yg, Arora2022-xd, Colin2022-dg, Kim2022-qc, Kim2023-vu, Doshi-Velez2017-qp}.
In our study, non-experts were asked to classify a set of images twice: first without the aid of the AI-identified words, and then again with the aid of the AI-identified words.
In the second session, the words were presented at the task-level as part of the instructions.
For instance, for the mass classification task, the following language was used: ``You will now be given words generated using an AI model to help describe general differences in the appearance of Benign vs. Malignant masses," followed by the listing of the top three scoring words for benign and malignant.
This level of instruction was minimal by design in order to assess whether the AI-identified words alone could help ``teach'' non-experts to perform the task above chance levels.
The participants were not given any other form of domain-specific training.
Participants were recruited who did not have a medical background and who were not affiliated with the project.
Out of the 12 total participants, 4 had post-graduate degrees (in non-medical domains), 2 had bachelor-level degrees, and 6 are currently still in training and do not currently have a college-level degree.
The study was performed for both the breast and skin tasks, with 11 and 12 participants completing the tasks respectively.
For each task, 50 images (25 benign, 25 malignant) were sampled from their respective test sets. Images were sampled randomly in a stratified manner such that the accuracy of the AI classifier on the 50 images approximately matched the accuracy across the entire test set.
This stratification was performed to obtain a representative sample in terms of task difficulty.

\subsubsection{Statistical Analysis}
Confidence intervals for area under the receiver operating characteristic curve (AUROC) were computed using the Delong method \citep{delong}.
Confidence intervals for accuracy were computed using the adjusted Wald method \citep{Agresti}.
Statistical comparison of human reader accuracy across sessions was performed using a one-sided paired t-test.
For assessing AI model accuracy, a default threshold of 0.5 was used to binarize scores.

\section{Results}

Our approach outlined in Figure 1 consists of training two linear models on top of a joint embedding space between images and text (i.e., CLIP \citep{CLIP}). The first model consists of a visual classifier trained to predict classification labels from image embeddings. The second model estimates the visual classifier as a linear combination of text embeddings. We performed experiments using two tasks: 1) benign vs. malignant classification of breast masses using the CBIS-DDSM dataset \citep{cbis}, and 2) benign vs. malignant classification of skin lesions using the SIIM-ISIC Melanoma Classification Challenge dataset \citep{siim-isic}. 


Consistent with prior linear-probe experiments using CLIP, we find that the linear visual classifiers perform relatively well on both the CBIS-DDSM and SIIM-ISIC datasets. On CBIS-DDSM, the model exhibits an area under the receiver operating characteristic curve (AUROC) of 0.715 (95\% CI: 0.661, 0.770) and accuracy of 64.4\% (95\% CI: 59.3\%, 69.1\%), which is similar to other reported performances \citep{Tu2023-ak}. On SIIM-ISIC, the performance is 0.831 AUROC (95\% CI: 0.784, 0.879) with an accuracy of 76.5\% (95\% CI: 71.2\%, 81.1\%), also in the realm of other performances considering no data augmentation or ensembling was used \citep{ADEPU2023}. Thus, even though CLIP was not specifically trained for medical imaging, the features it learned can facilitate both of these tasks, providing a basis for identifying words that reflect the classification tasks.

Figure 2 contains the weights for each word in estimating the visual classifiers. As described in the Methods, the list of words was created in a structured way to represent general-purpose visual properties. 
We find that the weights generally align with clinical intuition. 
For instance, for the breast mass task, the top weighted words for malignant are `asymmetric' and `large', and the top words for benign are `round' and `symmetric'. 
Round masses are indeed often benign and may represent a cyst or fibroadenoma, whereas malignant masses are often irregularly shaped \citep{breast_imaging}.
The top weighted words for the skin lesion task are `low contrast' and `coarse' for malignant and `smooth' and `round' for benign. 
Coarse versus smooth and round generally aligns with aspects of the ABCD criteria for skin lesions \citep{Friedman1985-zs}, where malignant lesions tend to have irregular borders and uneven colorization. 
However, the mapping to `low contrast' is less clear. 

\begin{figure}[h]
\floatconts
  {fig:word_weights}
  {\includegraphics[width=0.98\linewidth]{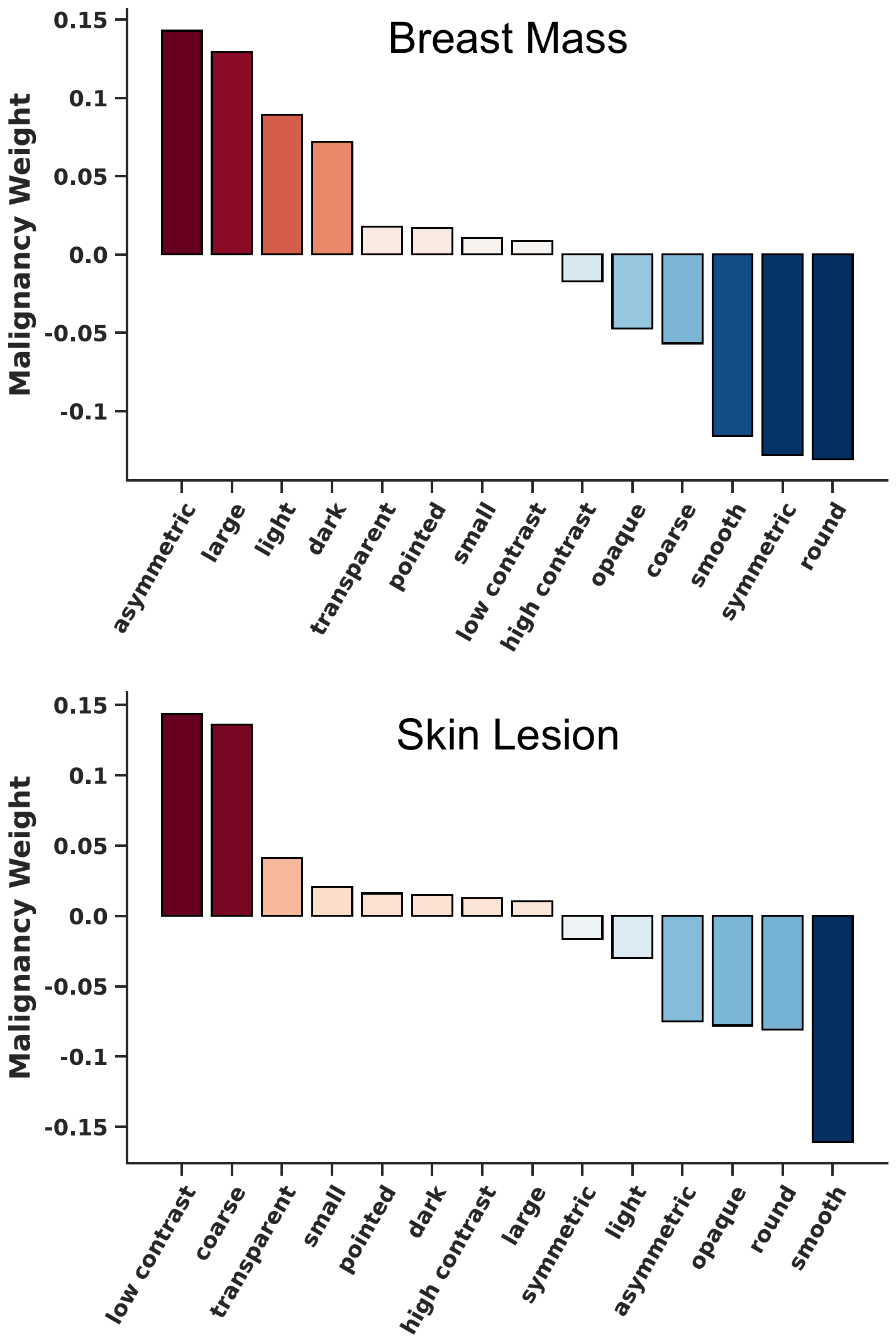}}
  {\caption{Word weights for each classification task.}}
\end{figure}
\vspace{-10pt}
\begin{figure*}[p]
\floatconts
  {fig:examples}
  {\centering}
  {\includegraphics[width=0.8\linewidth]{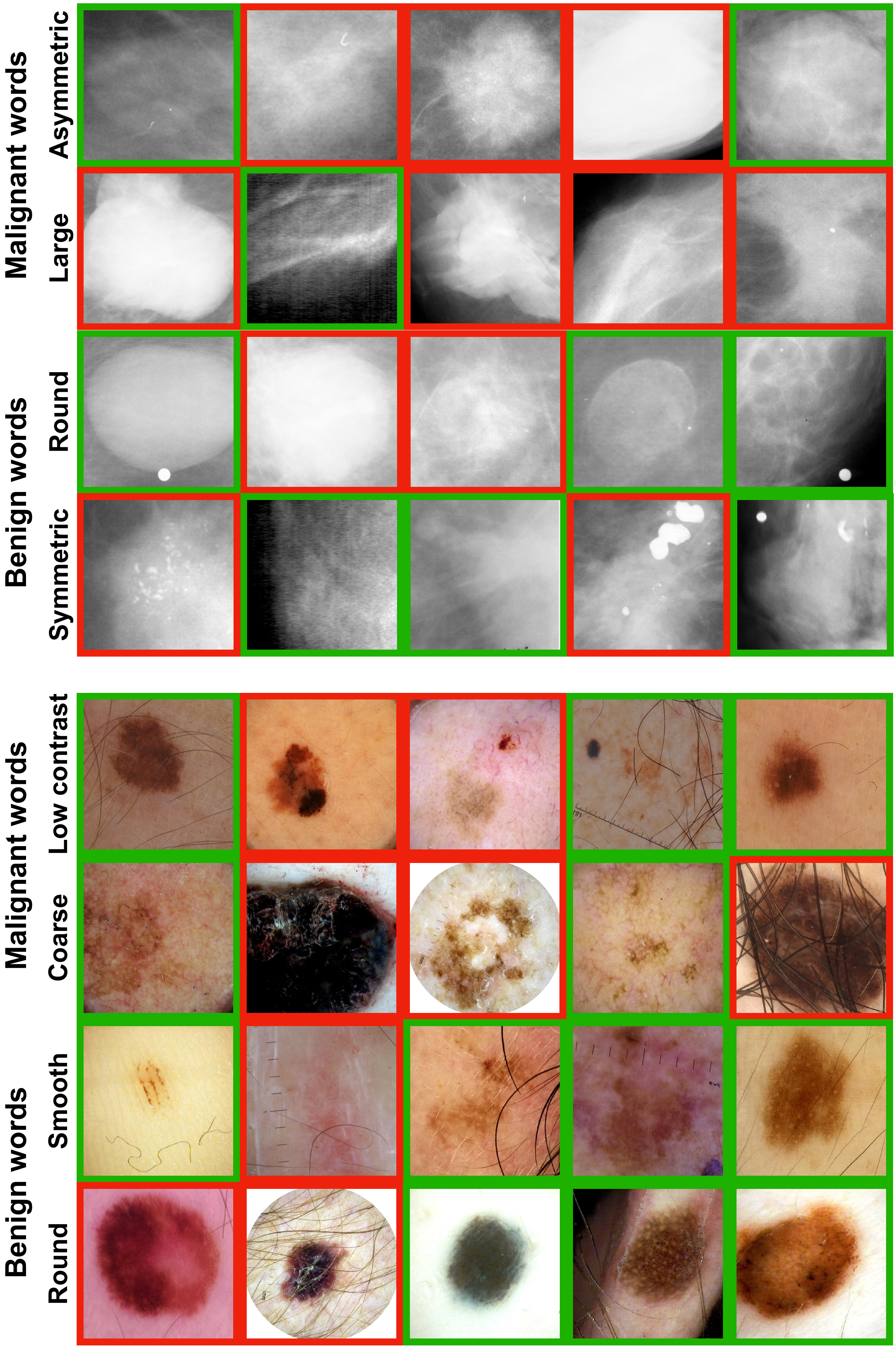}}
  {\vspace{-15pt}}
  {\caption{Prototypical examples for top weighted words for each task. Red border: malignant; green: benign.}}
\end{figure*}

To gain a better understanding of the features represented by the top weighted words, especially for `low contrast' in the melanoma task, we calculated prototypical examples for each word using the training datasets (see Methods). 
The top scoring prototypes are contained in Figure 3.
For words like `round', `smooth', and `coarse', the prototypical examples generally align with the intuitive meaning of these features. 
For `low contrast', this may reflect a less well-defined border between the lesion and the surrounding skin, which is also a concerning feature used by clinicians in diagnosing melanoma and aligns with the ABCD criteria \citep{Friedman1985-zs}.  
Thus, the prototypes are generally consistent with clinical considerations.
However, we note that not all word weights align with clinical knowledge, as `asymmetric' has a moderately negative (benign) weight for the melanoma task, whereas asymmetric lesions are generally more likely to be malignant.
Additionally, there remains much room to identify features that also correlate with the visual classifier, as the cosine similarity between the actual classifier weights and the estimated weights is 0.11 and 0.10 for the breast and melanoma tasks, respectively. 
One potential source for additional predictive features is shortcut connections, which we explore below. 

\subsection{Pilot Reader Study}
Beyond qualitative assessment of the word weights, we sought to more functionally assess explainability through a human reader study. 
Our rationale was as follows: if the top scoring words truly represent informative, human-interpretable features, then conveying these words to a novice human may enable them to perform the task above chance levels. 
We sought to test this in a minimal setting where participants are only provided explanations at the task-level via the list of top scoring words. 
We employ a two session design where the participants first perform the classification task without the AI-identified words and without any feedback. 
In the second session, the participants perform the same task again but are provided the top three AI-identified words for benign and malignant in the text instructions. 
Importantly, the same instructions apply to each image and the participants are not given any other form of AI assistance. 

Despite the highly-specialized nature of medical image interpretation, we do find some evidence that the AI-identified words can improve non-expert performance, particularly for melanoma. 
Before being provided the words for the melanoma task, the average accuracy of the 12 readers was 52.2\% (95\% CI: 48.2\%, 56.1\%), similar to the chance level of 50\%. 
After being provided the words, the average accuracy improved to 62.0\% (95\% CI: 58.1\%, 65.8\%) for an increase of of 9.8\% (p-value: 0.02; 95\% CI: 0.6\%, 19.1\%).
Figure 4 shows the change in performance for each participant, where 9 out of 12 participants improved and 3 participants had $\ge$70\% accuracy in the second session.
Thus, while the non-expert performance remained below expert and AI levels, simply providing the top AI-identified words enabled performance significantly above chance levels. 
For the breast mass task, reader performance was at chance in both sessions (Session 1: 48.4\% (95\% CI: 44.2\%, 52.5\%); Session 2: 52.4\% (48.2\%, 56.6\%).
We suspect that this may be due to the general difficulty of the task (where AI performance was also lower) and the lower quality of the images, which consist of ROIs from scanned film mammograms.

\begin{figure}[h]
\floatconts
  {fig:reader_study}
  {\caption{{\vspace{-18pt}}Reader study results for melanoma task.}}
  {\includegraphics[width=\linewidth]{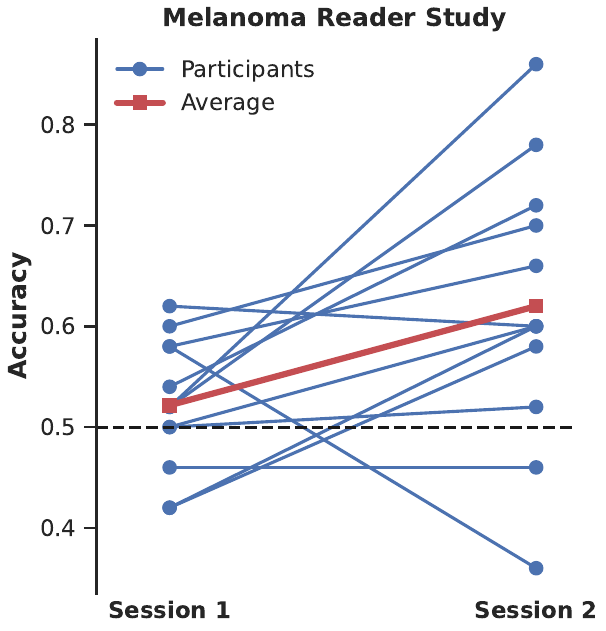}}
\end{figure}

While the pilot study provides a functional assessment of explainability, there are several inherent limitations.
For one, the objective of our study -- testing if the AI-identified words can enable non-experts to perform the task -- necessitates a sequential study design as the participants are no longer naive to the words after they are presented. 
This presents a possible confounder of readers getting better over time simply through experience, even if they are not provided with ground truth labels.
To explore this possibility, we computed the performance within each half of each session.
For session 1 in the melanoma task, readers demonstrated an average accuracy of 52.7\% in the first half of the session compared to 51.7\% in the second half.
For session 2, the average accuracy was 60.7\% in the first half of the session and 63.3\% in the second half.
Thus, the performance within sessions was much more similar than the performance across sessions, suggesting that `unsupervised' experience cannot solely explain the performance increase observed after presenting the AI-identified words.
We additionally performed subgroup analysis by reader education status to see if this factor can partially explain the results.
Participants without a college-level degree (n=6) demonstrated an average accuracy of 51.7\% in session 1 and 61\% in session 2 for the melanoma task.
Participants with a college-level degree (n=6) demonstrated similar performance, with accuracies of 52.7\% and 63\% in sessions 1 and 2, respectively.

\subsection{Exploring Shortcut Connections}
A purpose for which explainability is especially needed is dataset auditing, where AI models have been shown to learn `shortcut connections' between irrelevant confounders and task labels when such correlations exist in the training dataset \citep{Yan2023-gt, Kim2023-vt, Nauta2023-ph, Shirley_Wu2023-ac, Zhao2020-zc, De_Grave2023-sr}.
As there can be many possible confounders and manual assessment is infeasible in large datasets, approaches are needed to efficiently investigate such confounders.
We explored the potential of our approach in identifying shortcut connections by adding additional words to the initial word list. 
For the mass classification task, we included the word `clip' as metal clips known as biopsy markers are placed in the location of a biopsy for future reference. 
These clips have the potential to lead to shortcut connections for an AI model, especially if there is a correlation between the presence of clips and disease labels in the training dataset.
For the melanoma task, we use the word `marker' as suspicious lesions are often marked with an ink marker to aid in documentation and surgical planning.

\begin{figure}[h]
\floatconts
  {fig:shortcuts}
  {\caption{\vspace{-20pt}Prototypical examples for potential domain-specific shortcut connections.}}
  {\includegraphics[width=\linewidth]{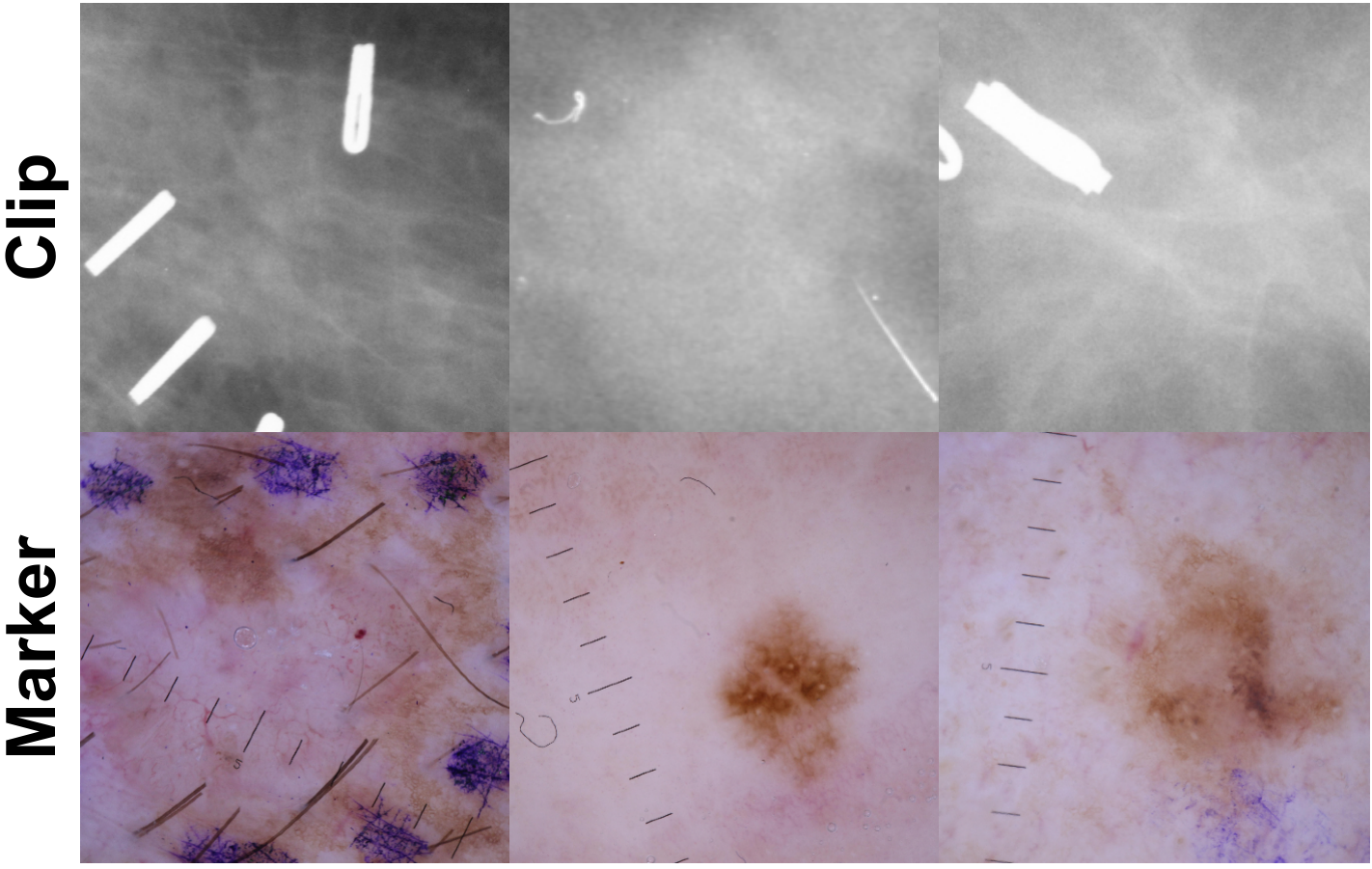}}
\end{figure}

Figure 5 contains examples of the top scoring prototypes for the considered shortcut words. The examples indeed illustrate the presence of the potential confounders, where, interestingly, the word `marker' may not only identify ink markings but also the markings of a dermoscopic ruler used to measure the lesion. 
However, the existence of these features within the datasets does not necessarily mean that their presence is correlated with the classification labels and/or an AI model would learn such correlations.
To assess this, we computed the probability of malignancy amongst images that score high for the shortcut word, as defined by scoring in the top 10\% of prototype scores. 
We additionally quantified the regression weight for the shortcut words in our classifier estimation approach.
For the word `clip' in the mammogram task, we find that the probability of malignancy is slightly higher in the top images for this word compared to the remainder of the images, at 57\% and 50\% respectively.
Consistent with this difference, the regression weight for this word is also slightly positive with a magnitude of 0.06.
For the word `marker' in the melanoma task, we find more dramatic correlations at both the dataset and model level.
The probability of malignancy is 45\% for the top `marker' images in the dataset, compared to 32\% for the remaining images.
This difference is reflected in a word weight of 0.11, which is of similar magnitude to the other top scoring words for malignancy.
Thus, without any data annotation, we find that our approach can efficiently identify and quantify shortcut connections, which can help guide strategies for mitigation and clinical use. 

\section{Discussion and Conclusions}
We presented a task-level, global explainability strategy that leverages a vision-language embedding space to identify descriptors of a visual classification task.
Through experiments using two medical imaging datasets, we find that the resulting top scoring words generally align with clinical knowledge and can be used to guide non-experts to perform a skin lesion task above a chance level.
We additionally show that our method can efficiently identify shortcut connections in the datasets used.
Importantly, we do not rely on extensive domain-specific language training or human supervision, which enables application to a variety of tasks and does not constrain the AI features used. 

There are several important future directions to build upon our results. 
For one, the described approach relies on a pre-trained vision-language model and thus cannot be straightforwardly adapted to vision-only classifiers.
Future work could learn a mapping from a vision-only representation space to a VLM embedding space to subsequently enable application of our technique.
Additionally, domain-specific VLMs and descriptors can be used to go beyond our general-purpose approach, which has the potential to more fully estimate the specialized classification tasks.
Altogether, more human-like explainability is a critical direction in AI research, where our approach represents a promising step towards intuitive, task-level explainability for visual tasks.  


\bibliography{ml4h}

\end{document}